\documentclass[10pt,twocolumn,letterpaper]{article}

\usepackage{cvpr}              %

\usepackage[dvipsnames]{xcolor}
\usepackage{multirow}

\definecolor{cvprblue}{rgb}{0.21,0.49,0.74}
\usepackage[pagebackref,breaklinks,colorlinks,citecolor=cvprblue]{hyperref}

\def\paperID{*****} %
\def\confName{CVPR}
\def\confYear{2024}

\title{Do Pre-trained Vision-Language Models Encode Object States?}

\author{Kaleb Newman \quad Shijie Wang \quad Yuan Zang \quad David Heffren \quad Chen Sun\\ \\
{Brown University}\\
}

\begin{document}
\maketitle
\begin{abstract}
For a vision-language model (VLM) to understand the physical world, such as cause and effect, a first step is to capture the temporal dynamics of the visual world, for example how the physical states of objects evolve over time (e.g. a \textit{whole apple} into a \textit{sliced apple}). Our paper aims to investigate if VLMs pre-trained on web-scale data learn to encode object states, which can be extracted with zero-shot text prompts. We curate an object state recognition dataset \textit{ChangeIt-Frames}, and evaluate nine open-source VLMs, including models trained with contrastive and generative objectives. We observe that while these state-of-the-art vision-language models can reliably perform object recognition, they consistently fail to accurately distinguish the objects' physical states. Through extensive experiments, we identify three areas for improvements for VLMs to better encode object states, namely the quality of object localization, the architecture to bind concepts to objects, and the objective to learn discriminative visual and language encoders on object states. Data and code are released at \url{github.com/brown-palm/object-states}.
\end{abstract}
    
\section{Introduction}
\label{sec:intro}

Vision-Language Models (VLMs) have become foundational in various visual understanding tasks, including object recognition~\cite{pmlr-v139-radford21a}, visual question answering~\cite{li2023blip}, and robotics applications~\cite{driess2023palm}. These models integrate visual and linguistic data, enabling nuanced interpretation and interaction with images and video. However, a critical yet underexplored aspect of VLMs is their ability to encode the physical states of objects—such as whether an apple is \textit{whole} or \textit{sliced}. Understanding these states is essential for physical commonsense reasoning, which underpins many practical applications, from assisting with daily tasks (e.g., recognizing that hot water can be poured into an empty glass) to enhancing interaction in robotic systems.

We define \textbf{object state} as the physical and functional condition or configuration of an object, as discernible from visual data. For example, an object's state can indicate whether it is melted, dirty, or undergoing a process such as cutting or pouring. While object state recognition is often associated with video analysis due to its applications in modeling temporal dynamics and action recognition, our study focused on temporally localized key frames that contain the object states of interest, from which human annotators are able to accurately label the states without temporal context, and on which pre-trained VLMs are applied to generate pseudo labels~\cite{xue2024learning}.

To explore VLMs' capabilities in this area, we employ them as zero-shot classifiers on a dataset we introduce called ChangeIt-Frames, based on~\cite{soucek2022lookforthechange}. This dataset contains images depicting various object states in natural scenes. We augment this dataset with bounding box annotations for a subset of 1,736 images, in which the original object state labels are verified by human annotators. We evaluate 9 state-of-the-art open-source models. We use two types of VLMs: dual-tower VLMs, which classify images based on the similarity between image and text embeddings based on contrastive learning, and Multimodal Large Language Models (MLLMs), which utilize a visual encoder paired with a generative language model backbone to respond to prompts. We observe that while these models excel in object recognition, they consistently underperform in reliably identifying the objects' physical states.

Recognizing object physical states in images presents unique challenges. Our study reveals that standard fine-tuning of VLMs with physically grounded data~\cite{dai2024instructblip}, or pre-training with curated datasets, does not necessarily enhance object state recognition performance on the ChangeIt-Frames dataset. We hypothesize that effective concept binding—linking visual features to corresponding objects—is crucial for VLMs to accurately discern object states. To test this hypothesis, we construct object-centric representations using CLIP and demonstrate that these modified VLMs improve at solving concept binding tasks involving color and shape, where vanilla CLIP struggles~\cite{lewis2022does}. Nonetheless, object-centric VLMs still exhibit limitations in recognizing object states, which we attribute to inadequate object localization and insufficiently discriminative visual and language representations. We additionally observe  an increase in model parameters or training data size can provide improvements but the performances are still far from satisfactory, and that the challenge prevails in Multimodal Large Language Models, such as LLaVA and PaliGemma.

\section{Recognizing Object States}\label{physical-states}

\noindent\textbf{Dataset:} We constructed our evaluation dataset, \textit{ChangeIt-Frames}, from the video-based ChangeIt dataset \cite{soucek2022lookforthechange}, which includes 650 videos of 44 object categories undergoing various state-changing actions. From these videos, we extracted 25,735 images, each depicting one of 96 distinct object states. This image-based dataset is exclusively used for zero-shot evaluation of Vision-Language Models (VLMs). To provide detailed annotations, we manually labeled a subset of 1,736 images from ChangeIt-Frames with bounding boxes around the target objects. Each image is labeled with a single bounding box corresponding to the object in its specific state. The annotation process was carried out using the default annotator pool from Amazon Mechanical Turk. These annotations are released under the MIT license. 

There have been introduced in the past to explore the compositionality of objects and their states, most notably MIT-States~\cite{StatesAndTransformations} and C-GQA~\cite{naeem2021learning}. Unlike C-GQA, which includes states (e.g., ``cute dog'') that may \textbf{not} necessarily be the result of observable physical changes, ChangeIt-Frames is exclusively concerned with irreversible physical changes in objects. This also separates ChangeIt-Frames from MIT-States, which organizes state variations primarily through adjective-noun pairings that may include reversible states like open/closed door, or the states of global objects like cluttered/empty room. Furthermore, MIT-States uses Bing
search engine with limited human annotation, leading to missing or inaccurate state labels.

\begin{figure}[h]
    \centering
    \includegraphics[width=\linewidth]{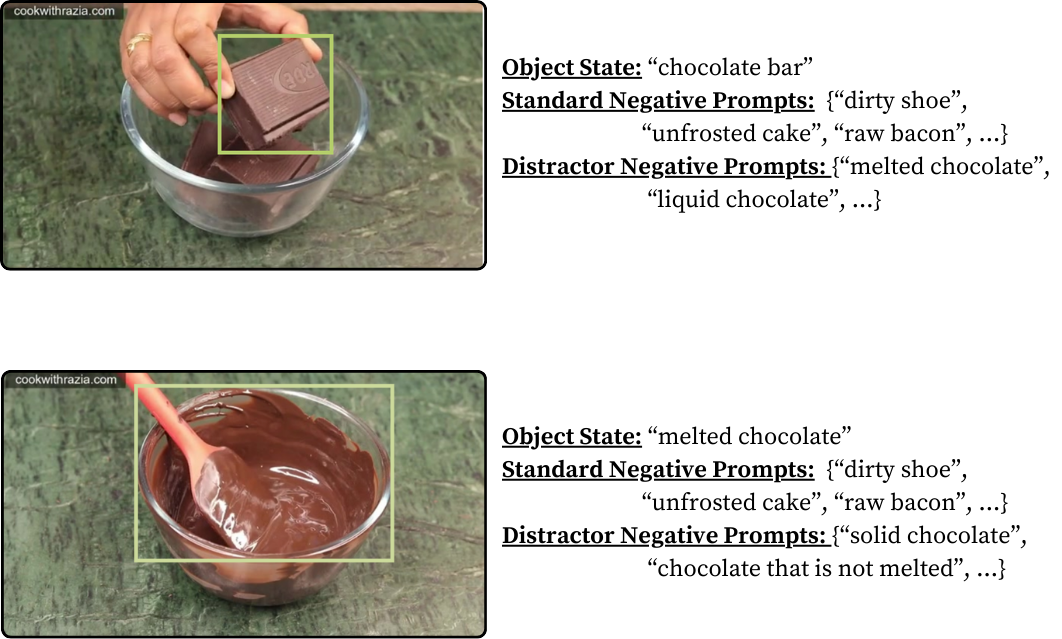}
    \caption{\textbf{ChangeIt-Frames dataset}. The images are sourced from instructional videos~\cite{soucek2022lookforthechange}. We use Amazon MTurk to manually verify a subset of the image annotations, and to draw bounding boxes for the objects of interest. For evaluation, we ask an VLM to choose the correct object state among ten candidates prompts selected via \textit{standard} or \textit{distractor} strategies.}
    \label{fig:change_it_frames}
\end{figure}

\noindent\textbf{Evaluation Setup:}
For each image, we select a list of descriptive prompts for possible object states, such as ``whole apple'' or ``fried bacon''. We refer to the correct description as a ``positive prompt'', and the incorrect descriptions as ``negative prompts''. We employ two strategies for selecting negative prompts: The \textbf{standard strategy} which selects the negative prompts corresponding to different states within the same object category, such as \textit{peeled apple} for \textit{whole apple}. The remaining negative prompts are randomly selected from the candidate pool to total 10 prompts per image. We also consider the \textbf{distractor strategy}, and the negative prompts consist of distractors specifically designed to be semantically similar yet incorrect regarding the object state. For example, for a positive prompt \textit{whole apple}, the distractor prompts might include \textit{an apple that is cut} or \textit{an apple that is peeled}. The remaining prompts are randomly selected, ensuring a total of 10 prompts. This setting is designed to challenge the model's ability to discern subtle differences in object states.

We utilize two methods for zero-shot classification based on the architecture of the Vision-Language Models (VLMs):
For the \textbf{dual-tower VLMs}, we compute the cosine similarity between the image and text embeddings. The label corresponding to the highest similarity is selected as the predicted output. For the \textbf{Multimodal Language Models} we present the models with a prompt formatted as: ``Which of these does this image depict: [numbered list of prompts]? Only reply with the single number that corresponds to the correct answer.'' The model's output is then used to determine the predicted label. These methods allow us to evaluate the models' ability to correctly identify object states across different architectures.

\noindent\textbf{Metrics:}
We separately calculate Object Accuracy and State Accuracy. For object accuracy, a prediction is considered correct if the predicted label includes the object's name (e.g., both \textit{whole apple} and \textit{cut apple} are correct for an image of an apple). For state accuracy, the model must predict the exact ground truth object state in the image.

\begin{table}[t]
  \centering
\begin{tabular}{ccccc}
\toprule
 & \multicolumn{2}{c}{\textbf{Standard}} & \multicolumn{2}{c}{\textbf{Distractor}} \\       \cmidrule(lr){2-3}
     \cmidrule(lr){4-5}
 \textbf{Model} & \textbf{Obj.} & \textbf{State} & \textbf{Obj.} & \textbf{State}\\ \hline
 CLIP & 0.918 & 0.614 & 0.925 & 0.408 \\
 OpenCLIP& 0.906 & 0.604 & 0.925 & 0.373 \\
 ALIGN & 0.910 & 0.620 & 0.918 & 0.403 \\
 FLAVA & 0.966 & 0.633 & 0.837 & 0.560 \\
 PhysVLM$^*$ & - & 0.338 & - & - \\\bottomrule
\end{tabular}
\caption{\textbf{Object and State Recognition} performance for both negative prompt sampling strategies illustrated in Figure~\ref{fig:change_it_frames}. Random performance is 10\%. $*$: Evaluated on cropped objects as recommended by the authors.}
\vspace{-1em}
\label{tab:standard_distractor_eval}
\end{table}

\noindent\textbf{Results and Analysis:}\label{std-eval}
We conduct experiments on CLIP ViT-L/14 \cite{pmlr-v139-radford21a}, OpenCLIP ViT-L/14 \cite{ilharco_gabriel_2021_5143773}, ALIGN \cite{alignmodel}, FLAVA \cite{singh2022flava}, and PhysVLM~\cite{gao2023physically}. Notably, CLIP, OpenCLIP, and ALIGN rely on image-level representation for image-text contrastive learning; FLAVA uses patch-level as opposed to image-level representations; PhysVLM fine-tunes InstructBLIP~\cite{dai2024instructblip} with ``physically grounded'' annotations collected on~\cite{zhu2023egoobjects}.
 Our results, summarized in Table~\ref{tab:standard_distractor_eval} , reveal that while object recognition accuracy is generally high, there is a consistent drop of approximately 30\% in state recognition accuracy. When distractor prompts are used, model performance generally drops significantly. FLAVA shows greater robustness with only a 7\% drop. Notably, incorrect predictions frequently correspond to our designed distractors.

\noindent\textbf{Discussion:} Our results show that simply fine-tuning with more physically grounded data, as in PhysVLM, does not help VLM better encode object states. 
It might even hurt the performance, presumably due to domain mismatch between fine-tuning data and ChangeIt-Frames. We further notice that FLAVA outperforms other VLMs under the more challenging distractor setup. We hypothesize that this may be due to its use of patch-level representation, which can support association of object regions and text descriptions. We test this hypothesis in the next section.

\section{Exploring Possible Remedies} \label{object-centric}

In this section, we explore potential solutions for improving the recognition of object states in Vision-Language Models. We hypothesize that VLMs fail to recognize physical states due to the lack of an explicit notion of \textit{objects} in these models. These models may process images as a ``bag of concepts'', associating them with the scene as a whole rather than with individual entities. 

To address this, we investigate the use of object-centric representations. We also evaluate the performance of larger VLMs trained on more extensive data to see if the low performance can be rectified by scale. The testbed for these improvements is also done on a subset of ChangeIt-Frames that include bounding box annotations and verified object state labels, both of which are done by human annotators.

To test whether focusing on specific objects can enhance state recognition, we implement object-centric VLMs. This approach involves isolating objects using bounding-box information, either provided by the dataset or generated by an off-the-shelf detection model like GroundingDINO~\cite{liu2023grounding}. By cropping the images to these object regions, we aim to create representations that explicitly associate visual concepts with distinct entities.

\subsection{Object-Centric VLMs}

We assess the effectiveness of object-centric representations in two main tasks: concept binding and physical state recognition. For concept binding, we use the CLEVR-Binding benchmark~\cite{lewis2022does}, which involves differentiating between visual concepts such as ``red cube and blue sphere'' versus ``blue cube and red sphere'', as well as spatial relationships of two objects, such as ``cube left of sphere'' versus ``sphere left of cube''. %
In Tables~\ref{tab:two_acc} and~\ref{tab:rel_acc}, we report accuracy on the training, validation, and generalization splits of CLEVR-Binding.
We observe that object-centric VLMs outperform image-level VLMs by huge margins on both tasks.

\begin{table}[h]
    \centering
    \resizebox{.95\linewidth}{!}{
    \begin{tabular}{lcccc}
    \toprule
    Model & Object-Centric & Train & Val & Gen \\
    \midrule
   CLIP & No & 27.02 & 7.17 & 31.40 \\
   NegCLIP & No & 25.39  &0.29  &41.33  \\\midrule
   CLIP& Yes & 93.96 &\textbf{94.12}  &\textbf{96.53}  \\
   NegCLIP &Yes &\textbf{96.58}  &71.20  &81.82  \\
    \bottomrule
    \end{tabular}
    }
    \caption{
    \textbf{Results on two-object adjective-noun binding task} introduced by CLEVR-Binding.
    }
    \vspace{-1em}
    \label{tab:two_acc}
\end{table}
\begin{table}[h]
    \centering
    \resizebox{.95\linewidth}{!}{
    \begin{tabular}{lcccc}
    \toprule
    Model & Object-Centric & Train & Val & Gen \\\midrule

   CLIP & No & 26.80 & 14.99 & 0.00 \\
   NegCLIP & No & 24.34  &1.57  &63.45  \\\midrule

CLIP & Yes &65.31  &54.48  &\textbf{92.48}  \\
NegCLIP& Yes &\textbf{89.51}  &\textbf{90.97}  &82.33  \\
    \bottomrule
    \end{tabular}
    }
    \caption{
    \textbf{Results for relational reasoning} in CLEVR-Binding.
    }
    \vspace{-.5em}
    \label{tab:rel_acc}
\end{table}

For physical state recognition, we evaluate VLMs on our human-annotated ChangeIt-Frames dataset subset using the full image, or ground truth objects. We present the results in Table~\ref{tab:golden_eval}. Although the results in Table~\ref{tab:two_acc} demonstrate that VLMs can effectively associate visual concepts with corresponding objects using object-centric representations, this improvement does not extend to better performance in recognizing physical states. This can be shown in Table~\ref{tab:golden_eval} (GT Crop) where models generally do not improve when provided with ground truth crops. This suggests that object crops do not compel models to do the fine-grained object analysis required to improve at state recognition.

\begin{table}[h]
  \centering
\begin{tabular}{ccccc}
\toprule
 & \multicolumn{2}{c}{\textbf{Standard}} & \multicolumn{2}{c}{\textbf{Distractor}} \\       \cmidrule(lr){2-3}
     \cmidrule(lr){4-5}
 \textbf{Model} & \textbf{Image} & \textbf{GT Crop} & \textbf{Image} & \textbf{GT Crop}\\ \hline
 CLIP & 0.655 & 0.642  & 0.463 & 0.465 \\
 OpenCLIP& 0.653 & 0.642  & 0.466 & 0.455 \\
 ALIGN & 0.715 & 0.702 & 0.502  & 0.548 \\
 FLAVA & 0.614 & 0.643 & \textbf{0.620} & \textbf{0.686}  \\
 \midrule
 ViT-G-14 & 0.710 & 0.726 & 0.494 & 0.520  \\
 SigLIP & \textbf{0.791} & \textbf{0.789} & 0.572 & 0.571 \\\bottomrule
\end{tabular}
\caption{\textbf{State Recognition} performance on our annotated ChangeIt-Frames subset. We show accuracy for whole image and ground truth crops under our Standard and Distractor setting. The last two models correspond to experiments run in Section~\ref{sec:larger}.}
\vspace{-1em}
\label{tab:golden_eval}
\end{table}

\subsection{Larger VLMs}\label{sec:larger}

We also investigate whether larger VLMs trained on extensive data can better recognize object states. Our evaluations include  OpenCLIP ViT-G-14 \cite{ilharco_gabriel_2021_5143773} and SigLIP \cite{zhai2023sigmoid}, we evaluate their performance on the standard and distractor settings of the annotated ChangeIt-Frames dataset. We observe that while larger dual-tower models show improved performance compared to CLIP and OpenCLIP, challenges remain under the distractor setting, where both are outperformed by FLAVA.

\subsection{Multimodal LLMs} 

In our state recognition experiments, we found that model performance typically declined in the distractor setting. To further explore state recognition, we examined Multimodal Large Language Models (MLLMs), a recent advancement over the Vision-Language Models (VLMs) previously used. Unlike VLMs, which rely on a standard text encoder, MLLMs incorporate a generative language model to process language inputs, significantly increasing the total model parameters. With this in mind, we tested whether the additional parameters and enhanced language capabilities of MLLMs would improve accuracy in the standard setting and address the more linguistically complex challenges posed by the distractor setting.

To investigate this, we asses the performance of PaliGemma~\cite{beyer2024paligemma} and two LLaVA-NeXT~\cite{liu2024llavanext} models (Mistral-7B and LLama-8B). The results in Table~\ref{tab:mllm_distractor_eval} show that the state recognition problem in dual-tower VLMs  translate to MLLMs. Even with the use of a Large Language Model (LLM) and extensive Visual Instruction Tuning, the distractor setting remains challenging.

\begin{table}[h]
  \centering
\begin{tabular}{lccc}
\toprule
 \textbf{Model} & \textbf{Standard} & \textbf{Distract} \\
 \hline
 PaliGemma & 0.716 & 0.242 & \\ 
 LLaVA-Mistral-7B & 0.767  & 0.656 & \\ 
 LLaVA-Llama-8B& 0.579 & 0.430 & \\\bottomrule

\end{tabular}
\caption{\textbf{State Recognition} performance on the annotated subset for selected MLLMs. The last two models are both versions of LLaVA-NeXT. Each column corresponds to the different evaluation settings, detailed in Section~\ref{physical-states}.}
\vspace{-1em}
\label{tab:mllm_distractor_eval}
\end{table}

\section{Inspecting the Encoded Representations}\label{encoding}

As we have ruled out several likely remedies to fix existing pre-trained VLMs on recognizing physical states of objects, we now aim to investigate why such models fail by inspecting their encoded visual and text representations.

\begin{figure}[h!]
    \centering
    \includegraphics[width=\linewidth]{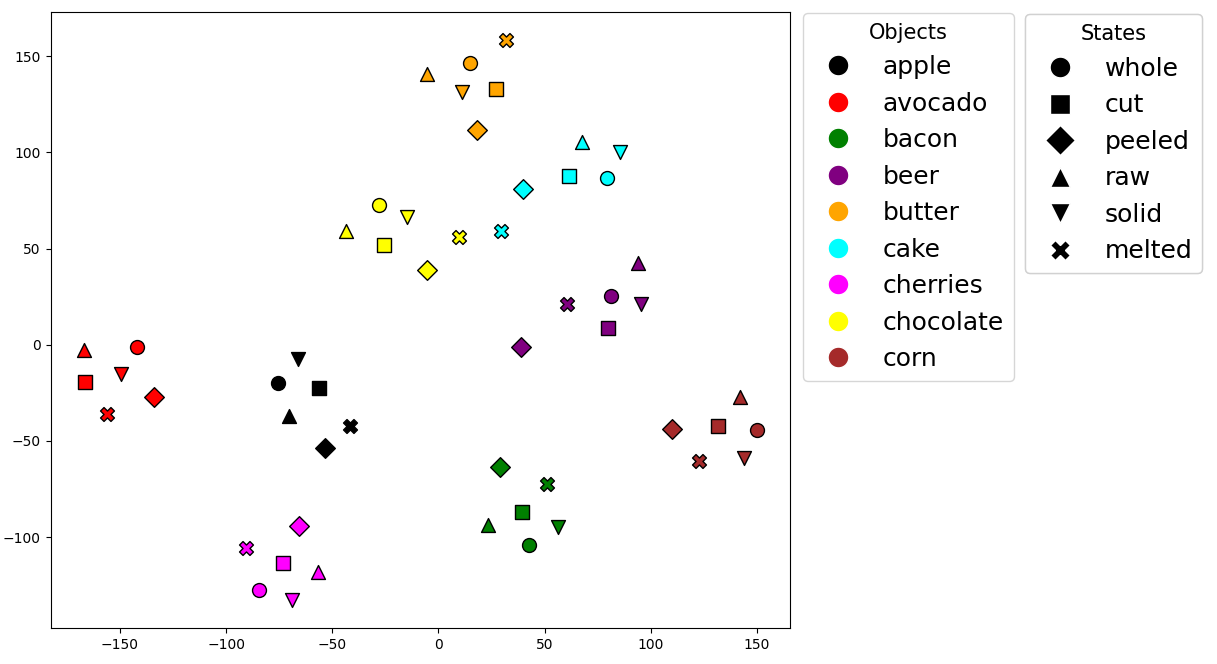}
    \caption{The T-SNE visualization of CLIP text embeddings. The representations of the text prompts are clustered by the object category and the representations for different states of the same object are very similar.}
    \label{fig:text_tsne}
\end{figure}

We first investigate whether the text encoder can properly reflect the physical state descriptions, we utilize T-SNE to visualize the CLIP text embedding of different text prompts of ``state + object'' combinations. As illustrated in Figure~\ref{fig:text_tsne}, the representations of the text prompts are clustered by the object category rather than the physical state, this indicates that the text encoders fail to learn discriminative representations for physical states of objects.

\begin{figure}[h]
  \centering
  \includegraphics[width=0.42\textwidth]{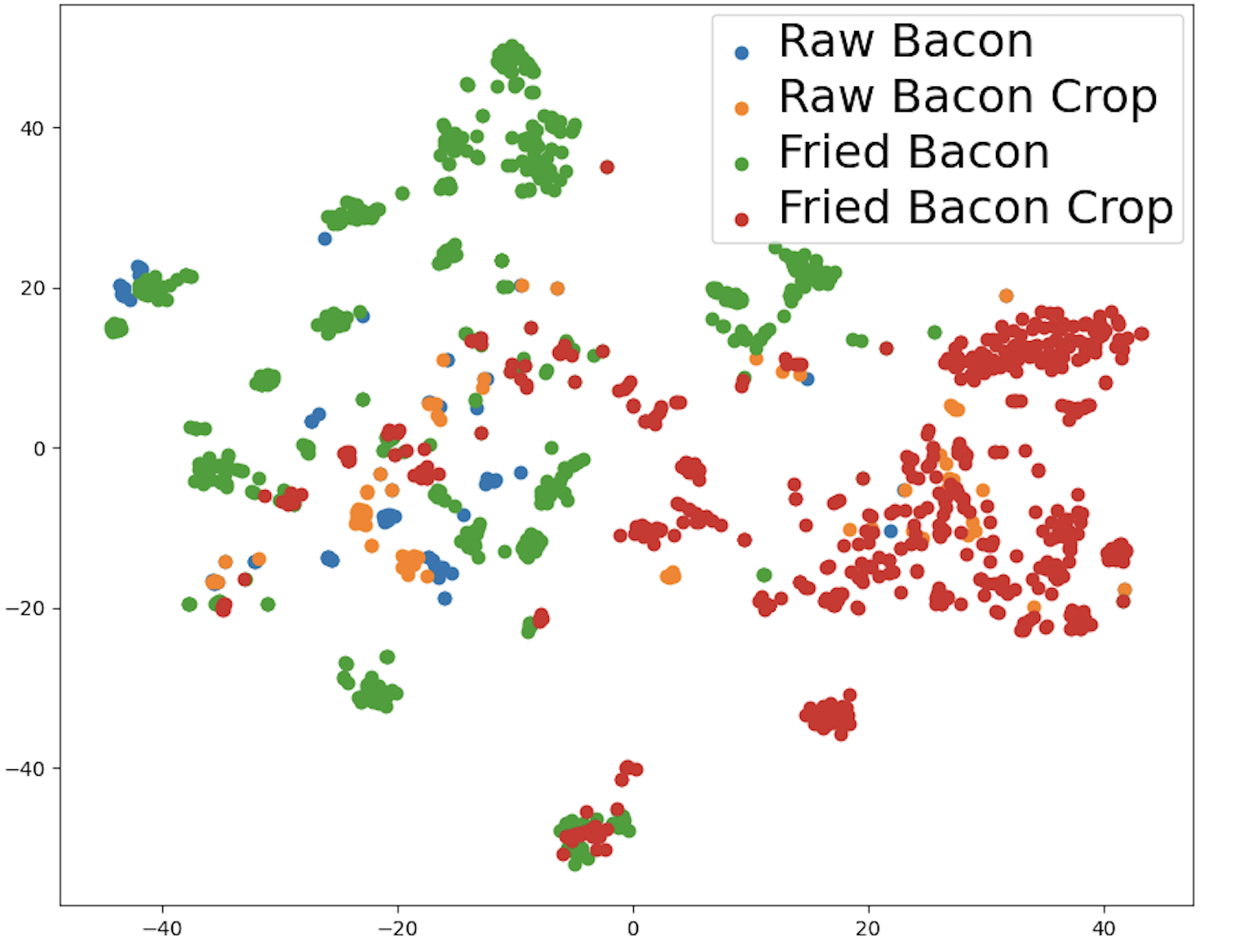}
  \caption{TSNE projections of CLIP visual embeddings for ``bacon''. This includes original and cropped images for both states.} 
  \vspace{-1em}
  \label{fig:bacon-tsne}
\end{figure}

We further validate the lower performance by visualizing the distributions of encoded object-level and image-level visual representations for the same objects with opposite states. We observe that the t-SNE projections do not show a clear distinction between the states, nor are the cropped images embeddings for a given state closer to the  whole image embedding.  We also observe that cropping has a larger effect on the representation than the state itself, suggesting the embeddings are not robust to transformation. An example of this can be seen in Figure~\ref{fig:bacon-tsne}.

Although we have demonstrated the lack of discriminative information in the encoded text and visual representations for object states, there are many plausible remedies that need to be investigated. We hypothesize that a combination of stronger object localization and modeling in the VLMs, combined with training objectives that explicitly encourage the recognition of object states (e.g. tracking objects that may undergo state transformations in video data), would be needed to encourage VLMs to capture object physical states.

\section{Conclusion}
Despite excellent performance on zero-shot object recognition, we demonstrate that existing pre-trained Vision-Language Models struggle to encode object state information, which we believe hinder their capabilities to understand and reason about the physical world. We hypothesize the challenge may come from lack of physically grounded training data, or the lack of object-centric inductive bias for VLMs to bind concepts to objects. We collect the ChangeIt-Frames benchmark with object bounding box and physical state annotations, and conduct extensive evaluations. We observe that addressing the data or model architecture issues alone does not solve object state recognition, and expect further progress to be made on object localization quality, concept binding, and pre-training objectives. We hope our findings will help develop future generation VLMs that can better capture object states.

\noindent\textbf{Limitations:}
Our evaluation mainly relies on a single dataset derived from instructional videos collected from the internet. Although the annotations are collectively manually, they are still subject to label noise. Evaluation performed on more diverse visual domains is desired.

\noindent\textbf{Acknowledgements:}
We would like to thank Nihal Nayak for help with CLEVR-Binding, and Licheng Yu, Tian Yun, and Ning Zhang for valuable feedback. The project was in part supported by Meta AI.

{
    \small
    \bibliographystyle{ieeenat_fullname}
    \bibliography{states}
}

\setcounter{table}{0}
\renewcommand{\thetable}{A\arabic{table}}
\setcounter{figure}{0}
\renewcommand{\thefigure}{A\arabic{figure}}

\appendix

\section{Appendix}
\label{sec:appendix}
\subsection{Model Descriptions}\label{sec:model_desc}
\textbf{CLIP/OpenCLIP ViT-L/14} The first models we consider are CLIP \cite{pmlr-v139-radford21a} and OpenCLIP \cite{ilharco_gabriel_2021_5143773} which share the same architecture and training. The main difference is that CLIP is trained with the private WebImageText dataset whereas OpenCLIP is trained with the public LAION dataset. For both CLIP and OpenCLIP model we use the ViT-L/14 architecture. For our analysis we calculate the cosine similarity between the encoded image and encoded text - further described in. \\
\textbf{ALIGN} We next use the ALIGN \cite{alignmodel} which is proposed to leverage a noisy dataset of over one billion image alt-text pairs. The ALIGN is a dual encoder with EfficientNet as its vision encoder and BERT as its text encoder.\\
\textbf{FLAVA} 
FLAVA is optimized on multiple vision, language, and cross- and multi-modal tasks \cite{singh2022flava}. The FLAVA model contains an image encoder and a text encoders well as a multimodal encoder combines image and text representations for higher-level tasks. Both image and text encoder follows a ViT-B/16 architecture that outputs a list of state vectors. For the multimodal encoding, the two lists of state vectors are further forwarded to a transformer module that is based on the ViT architecture. Different from the other models, FLAVA learns representations from multimodal data (image-text pairs) and unimodal data (unpaired images and text). For our experiments, we use  the image and text encoders and compute text-image similarity in a 768-dimensional embedding space. \\
\textbf{PhysVLM}
We use the PhysVLM model~\cite{gao2023physically}, fine tuned on the PhysObjects dataset, to evaluate how a model grounded in the physical world, focused on understanding physical concepts performs on this sort of task. Intuitively, understanding of physical reasoning would help differentiating object states. Specifically, we use PG-InstructBLIP, a fine-tuned version of InstructBLIP with the language model Flan-T5-XXL. We use the prompt "Question: Does this frame depict 1. {init state text}, 2. {action text}, 3. {end state text} or 4. none of the above" and query each video frame to generate predictions and confidence scores for each. 
\textbf{GroundingDINO} GroundingDINO is a zero-shot object detection model that can identify objects based on textual input \cite{liu2023grounding}. GroundingDINO combines a Transformer-based DINO detector with grounded pre-training. We use this model to provide object-centric information to our experiments. \\

\begin{figure}[h]
  \centering
  \includegraphics[width=0.5\textwidth]{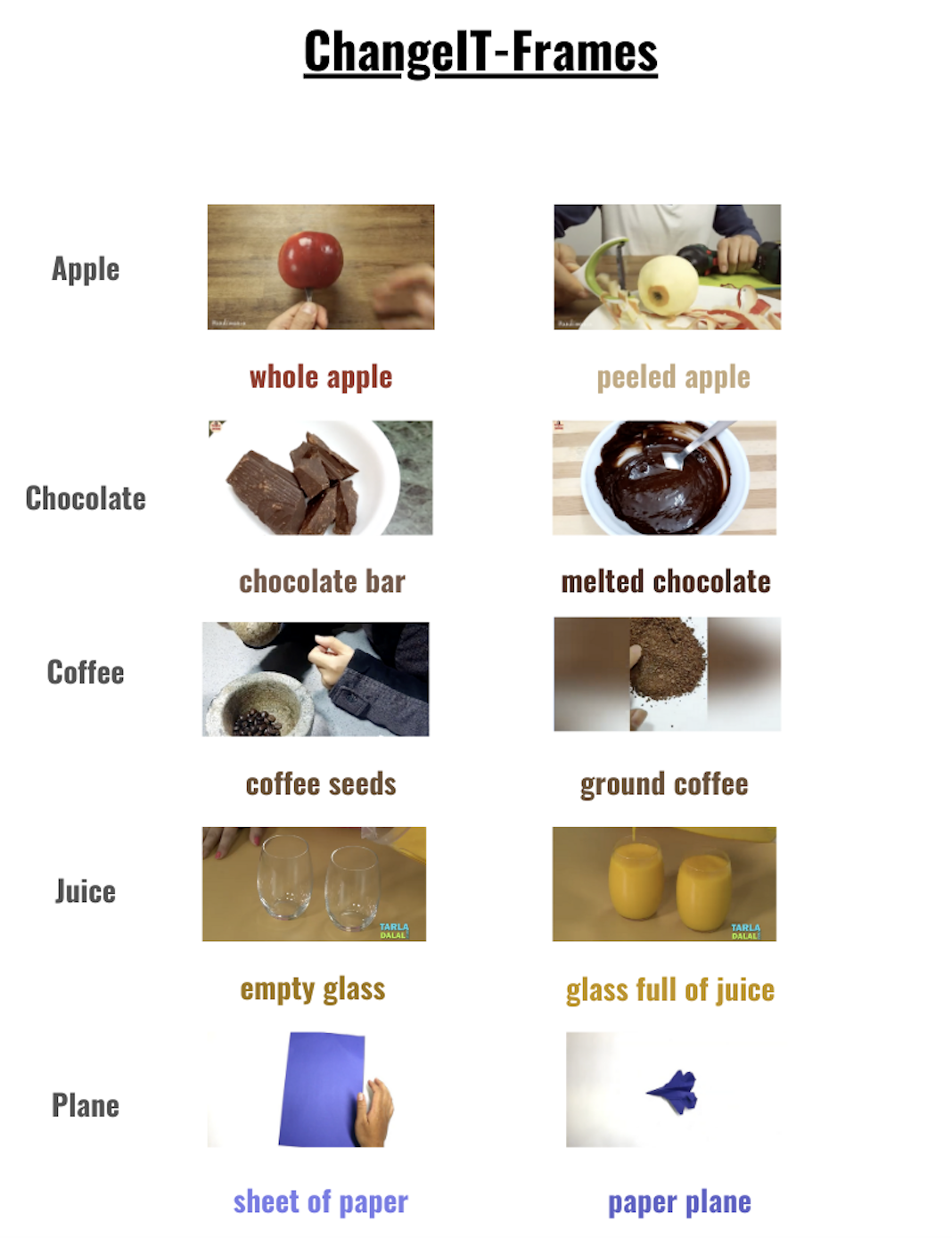}
  \caption{Visualization of Initial and Terminal states from categories in \textbf{ChangeIT-Frames}} 
  \label{fig:changeit_overall}
\end{figure}

\subsection{Data Collection Methodology}
For data annotation, we provided clear and concise instructions to the participants involved in the image annotation task. The instructions were as follows:
``Please draw one bounding box around the object matching one of the provided attribute labels. If the object is unclear or obscured by text, use 'Nothing to Label'. Draw only ONE bounding box per image. If multiple objects with the same attribute are close together, include all in a single bounding box. If objects are in different states making it difficult to identify them, label the image as 'Cannot Determine'.''
Additionally, to ensure clarity and accuracy in the annotations, we provided examples of both good and bad annotations through screenshots. This was intended to guide the annotators in making judicious decisions about the objects and their states in the images.
Participants were recruited via Amazon Mechanical Turk, a widely used crowdsourcing platform, and the labeling tasks were set up using Amazon SageMaker. We recruited participants under the default setting. The payment for the task was determined according to Amazon's recommended guidelines, ensuring fair compensation based on the complexity of the task and the typical rates in the participants' countries of residence. This approach helped us maintain ethical standards while also attracting competent and motivated annotators.

\subsection{Details on CLEVR-Binding}

The CLEVR-binding dataset is split into training, validation, and generalization set with distinct attribute-noun pairs. For each object in each image, the answer candidates are formed by the ground-truth pair and four distractor pairs. In details, in the example of image with a \textit{red sphere} and \textit{blue cube}, the answer candidates for \textit{red sphere} are composed of itself and two distractors that switching the existing attribute and noun compositions (\textit{red cube} and \textit{blue sphere}) and two randomly sampled from other negative pairs.
To zero-shot evaluate CLIP and its variants' ability of attribute-noun binding, we design the following prompt recommended by OpenAI: ``a photo of [adj] [noun]''.

\subsubsection{Relational Reasoning}

The two-object relational reasoning task requires models to predict the objects in the scene and their relationship. The CLEVR-binding dataset contains 3 types of objects: $\{$cube, sphere, cylinder$\}$ and 4 types of relationship: $\{$left, right, front, behind$\}$ with 24 possible combinations of spatial relations (Note that the relations are “symmetric”, e.g. cube left sphere is equivalent to sphere right cube).

In the setting of relational reasoning, an input image $v$ containing two objects $(s, o)$. The goal is to predict the relationship in the format of subject relation-object triple $(n_s, R, n_o)$ where $n_s, n_o \in \mathbb{N}=\{$cube, sphere, cylinder$\}$ is the shape of object $s$ and $o$ and $R\in\mathbb{R}=\{$left, right, front, behind$\}$ is the spatial relationship.

We propose a two-stage method for relational reasoning. In the first stage, CLIP takes in the object-centric representations $(x_s, x_o)$ extracted by the frozen CLIP vision encoder from the masked images, and zero-shot recognize the objects in the image, where  $n \in \mathbb{N}$ and $\mathcal{T}$ is the frozen CLIP text encoder:
\begin{align}
n_s &= \arg \max_{n} (x_s \cdot \mathcal{T}(\texttt{a photo of } n )) \\
n_o &= \arg \max_{n} (x_o \cdot \mathcal{T}(\texttt{a photo of } n ))
\end{align}
In the second stage, a linear head $\mathcal{L}$ takes in the object-centric representations to predict the relation $R$ between object $s$ and $o$. The linear head is trained on the training set.
\begin{align}
R &= \mathcal{L}(x_s\Vert x_o)
\end{align}

\end{document}